%% file: main.tex
\documentclass[letterpaper, 10 pt, conference]{ieeeconf}  % Comment this line out if you need a4paper
\usepackage[dvipsnames]{xcolor}
\usepackage{subfig}
\IEEEoverridecommandlockouts % This command is only needed if you want to use the \thanks command
\input{commands}

\title{\LARGE \bf
The Impact of 2D Segmentation Backbones on Point Cloud Predictions Using 4D Radar
}
\author{William L. Muckelroy III$^{1,3}$, Mohammed Alsakabi$^2$, John M. Dolan$^3$, Ozan K. Tonguz$^2$
\thanks{W. L. Muckelroy III is with the Robotics Institute, Carnegie Mellon University, Pittsburgh, PA, USA, and the Electrical \& Computer Engineering Department, University of Pittsburgh, Pittsburgh, PA, USA.
Email: \href{mailto:wlm14@pitt.edu}{\tt wlm14@pitt.edu}, \href{wmuckelr@andrew.cmu.edu}{\tt wmuckelr@andrew.cmu.edu}} 
\thanks{M. Alsakabi is with the Department of Electrical \& Computer Engineering, Carnegie Mellon University, Pittsburgh, PA, USA. Email: \href{mailto:malsakabi@cmu.edu}{\tt malsakabi@cmu.edu}}%
\thanks{J. M. Dolan is with the Robotics Institute, Carnegie Mellon University, Pittsburgh, PA,
USA. Email: \href{mailto:jdolan@andrew.cmu.edu}{\tt jdolan@andrew.cmu.edu}}
\thanks{O. K. Tonguz is with the Department of Electrical \& Computer Engineering, Carnegie Mellon University, Pittsburgh, PA, USA. Email: \href{mailto:tonguz@andrew.cmu.edu}{\tt tonguz@andrew.cmu.edu}}
}

\usepackage{siunitx}
\usepackage{multirow} % For multi-row cells
\usepackage{tabularx}
\newcolumntype{C}{>{\centering\arraybackslash}X}
\usepackage{caption}
\usepackage[ruled,vlined]{algorithm2e}
\DontPrintSemicolon  % Uncomment this if you don't want semicolons at the end of each line

\usepackage{algpseudocode}

\usepackage{amsmath}    % For math symbols
\usepackage{amssymb}    % For additional math symbols | % For \checkmark http://ctan.org/pkg/amssymb
\usepackage{mathtools}  % For more math tools
\usepackage{bm}         % For bold math symbols

\usepackage{hyperref}   % For references
\usepackage{accents}  % Add this in the preamble

\usepackage{graphicx}  % For including emojis
\usepackage{colortbl}  % For table colors
\usepackage{array}     % For custom column width
\usepackage{adjustbox} % To fit table in text width
\usepackage{pifont}  % For \ding{55} (X mark) http://ctan.org/pkg/pifont
\usepackage{color}  % For colored circles
\usepackage{subfig}  % Use subfig instead of subcaption for IEEE format

\begin{document}

\maketitle
\thispagestyle{empty}
\pagestyle{empty}

%%%%%%%%%%%%%%%%%%%%%%%%%%%%%%%%%%%% ABSTRACT %%%%%%%%%%%%%%%%%%%%%%%%%%%%%%%%%%%%
\begin{abstract}
LiDAR's dense, sharp point cloud (PC) representations of the surrounding environment enable accurate perception and significantly improve road safety by offering greater scene awareness and understanding. However, LiDAR's high cost continues to restrict the broad adoption of high-level Autonomous Driving (AD) systems in commercially available vehicles. Prior research has shown progress towards circumventing the need for LiDAR by training a neural network, using LiDAR point clouds as ground truth (GT), to produce LiDAR-like 3D point clouds using only 4D Radars. One of the best examples is a neural network created to train a more efficient radar target detector with a modular 2D convolutional neural network (CNN) backbone and a temporal coherence network at its core that uses the RaDelft dataset for training \cite{roldan_deep_2024}. In this work, we investigate the impact of higher-capacity segmentation backbones on the quality of the produced point clouds. Our results show that while very high-capacity models may actually hurt performance, an optimal segmentation backbone can provide a 23.7\% improvement over the state-of-the-art (SOTA).
\end{abstract}
%%%%%%%%%%%%%%%%%%%%%%%%%%%%%%%%%%%% INTRODUCTION %%%%%%%%%%%%%%%%%%%%%%%%%%%%%%%%%%%%
\section{Introduction}
\label{sec:intro}

There are numerous reasons why widespread adoption of commercially available autonomous vehicles (AV) and commercially viable autonomous driving (AD) has not occurred, one of which is the cost-prohibitive sensor suites they employ. These systems are designed to significantly improve road safety through their greater awareness and understanding of the scene, allowing them to interact with the world and react to environmental disturbances in a safe manner. However, an essential component of these systems is LiDAR because of the sharp, dense point cloud (PC) representations of the environment they can produce. These PCs are often used with arrays of radars and cameras to detect people, cars, and other objects as the ego vehicle traverses the world. Of these three primary sensors, we see that LiDARs take a significant stake in the overall cost of the sensor suite in AVs, ranging from \$4,000 for mid-range LiDARs to \$70,000 for long-range high-end LiDARs like that used in the KITTI dataset \cite{collin_resilient_2019}\cite{geiger_vision_2013}. However, 4D Radar is a relatively new emerging sensor in the AD space that offers promising performance at lower costs, as seen in \autoref{tab:sensors_comparison}.

\begin{table*}[ht]
\centering
\caption{Comparison of common Autonomous Driving sensors and data formats \cite{han_4d_nodate}\cite{fan_4d_2024}.}
\resizebox{\textwidth}{!}{%
\begin{tabular}{lccccc}
\hline
\textbf{Features} & \textbf{4D mmWave Radar} & \textbf{3D mmWave Radar} & \textbf{LiDAR} & \textbf{RGB Camera} & \textbf{Thermal Camera} \\
\hline
Range Resolution         & High     & High         & Very High    & Low          & Low \\
Azimuth Resolution       & High     & Moderate     & Very High    & Moderate     & Moderate \\
Elevation Resolution     & High     & Unmeasurable & Very High    & Moderate     & Moderate \\
Velocity Resolution      & High     & High         & Unmeasurable & Unmeasurable & Unmeasurable \\
Detection Range          & High     & High         & Moderate     & Low          & Moderate \\
Surface Measurement      & Texture  & Texture      & None         & Color        & Thermal Signature \\
Lighting Robustness      & High     & High         & High         & Low          & High \\
Weather Robustness       & High     & High         & Low          & Low          & High \\
Cost                     & Moderate & Low          & High         & Moderate     & High \\
\hline
\end{tabular}%
}
\label{tab:sensors_comparison}
\end{table*}

Previous research has shown that one can build a system that has the potential to replace LiDAR by leveraging concepts from Deep Learning (DL). By training a neural network using LiDAR PCs as ground truth (GT), one can produce LiDAR-like 3D PCs using only 4D radars \cite{roldan_deep_2024}\cite{roldan_see_2024}\cite{brodeski_deep_2019}\cite{cheng_novel_2022}\cite{fan_4d_2024}\cite{han_4d_nodate}. An example of such a network is the neural network proposed in \cite{roldan_deep_2024} and trained on the RaDelft dataset as a more efficient radar target detector. Using a modular and straightforward 2D convolutional neural network (CNN) backbone for segmentation and a temporal coherence network enabled the authors of \cite{roldan_deep_2024} to achieve significant improvements in bidirectional chamfer distance (BCD) when compared to other methods like various forms of Constant False Alarm Rate (CFAR) (i.e., CA-CFAR, SOCA-CFAR, GOCA-CFAR, OS-CFAR, ML-CFAR, etc.) \cite{jimenez_jimenez_general_2022}\cite{hatem_comparative_2018}\cite{jalil_analysis_2016}.

\begin{figure}
    \centering
    \includegraphics[width=1\linewidth]{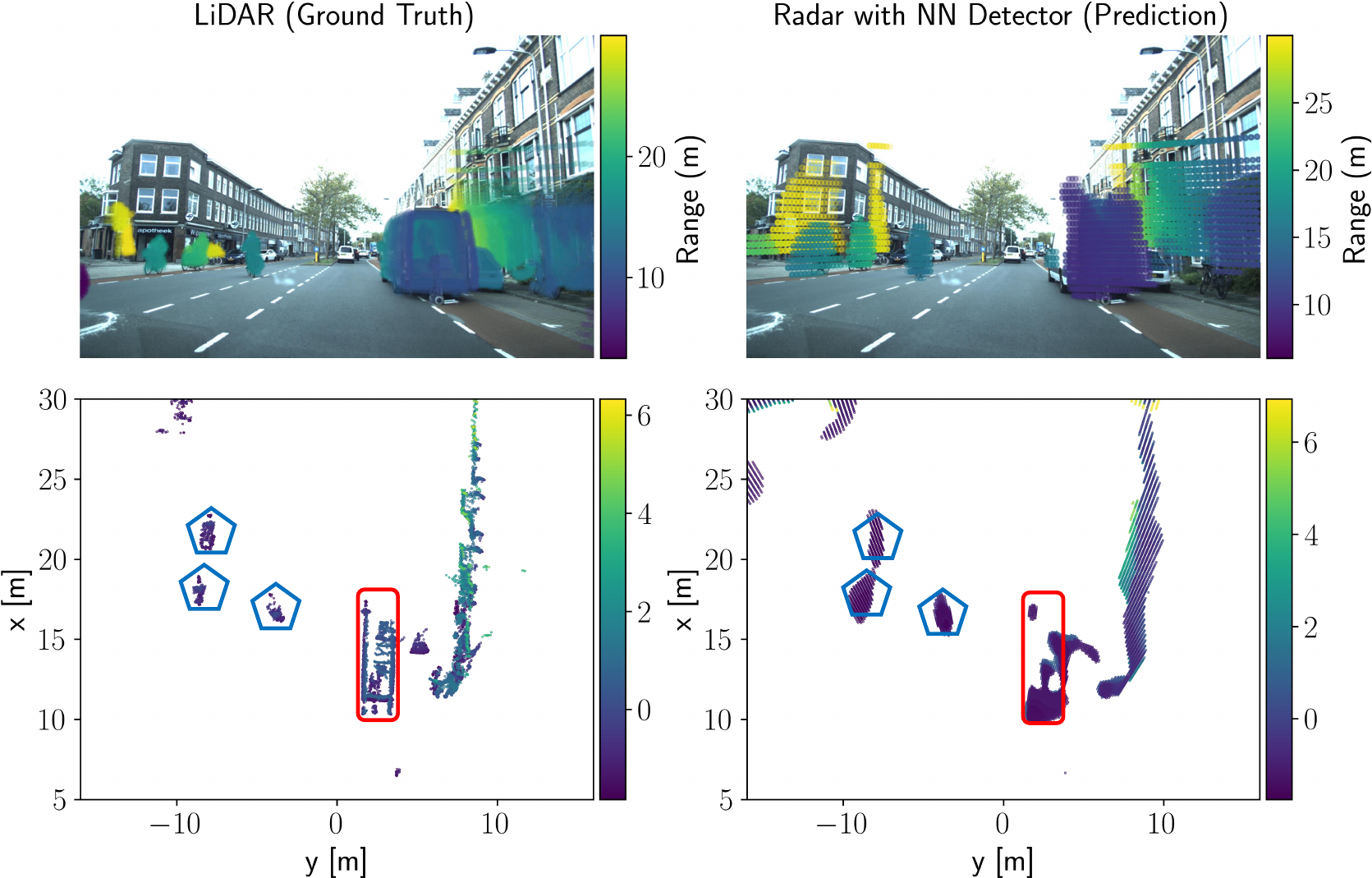}
    \caption{Example LiDAR ground truth and predicted 4D radar PCs. Left, a projection of the LiDAR ground truth point cloud into the camera and a BEV (bird's-eye-view). Right, a projection of the Radar + NN (neural network) predicted point cloud into the camera and its corresponding BEV.}
    \label{fig:delft-lidar-cfar-example1}
    \vspace{-5 mm}
\end{figure}

Our work further investigates the impact of the 2D segmentation backbone and the number of 3D convolutional layers in the temporal coherence network on the overall performance of the predicted point clouds. This allows us to gain insight into how optimizations of the overall architecture can further minimize BCD and increase the viability of this technique as a potential replacement for LiDAR.

The remainder of this paper is organized as follows. \autoref{sec:background} presents the background of the problem investigated and different definitions of key terms used throughout the paper. \autoref{sec:methods} outlines the methods used in this work for experimentation and how we evaluate the performance of such experiments. \autoref{sec:results} presents the primary findings of our work. Finally, \autoref{sec:conclusion} summarizes the discussions that have been held throughout this paper, along with our intentions to further improve the metrics and results in the future.

%%%%%%%%%%%%%%%%%%%%%%%%%%%%%%%%%%%% BACKGROUND %%%%%%%%%%%%%%%%%%%%%%%%%%%%%%%%%%%%

\section{Background}
\label{sec:background}

\subsection{A Backstory on 4D mmWave Radar}
\label{sec:4d-radar-history}

4D mmWave radar is a relatively new technology that provides useful Doppler and elevation information at relatively low cost, leading to multiple studies in data generation, perception, and SLAM (Simultaneous Localization and Mapping) \cite{han_4d_nodate}\cite{fan_4d_2024}. However, it has some characteristics that have created significant barriers to its widespread adoption. First, the 4D radar generates significantly more raw data than its 3D counterpart, which poses a problem with respect to optimal signal processing for resource-constrained applications. Second, PC data generated by the 4D radar are significantly noisier and sparser than LiDAR in traditional signal processing workflows. This noise and sparsity make it difficult for catered algorithms like CFAR to perform well in diverse scenarios and produce dense scene representations needed for AD applications.

Analyzing advances in 4D radar data generation over the last few years, we observe two common trends using DL concepts to improve the fidelity of 4D radar imaging: reconstruction and detection \cite{han_4d_nodate}. 

\subsubsection{Reconstruction}
\label{sec:reconstruction}

Techniques attempting to use DL models to improve the density and resolution of the typically sparse point clouds produced by traditional filtering methods like CFAR. PC reconstruction has been a common solution to improve the spatial resolution of LiDAR PCs with rooted examples such as PointNet, PCN, and R2P that demonstrate promising results \cite{qi_pointnet_2017}\cite{yuan_pcn_2019}\cite{sun_r2p_2022}. However, there are still shortcomings in reconstruction in AD applications. For example, reconstruction methods such as \cite{sun_3drimr_2021} require multiple viewpoints and poses for the model to reconstruct the target accurately. As most sensors in AD applications are statically mounted to a vehicle, it remains infeasible for reconstruction models to perform optimally without additional variability in viewpoint angle unless the vehicle is rotating around a given target.

\subsubsection{Detection}
\label{sec:detection}

These methods process raw frequency data directly, allowing the retention of information that is often lost with filtering methods like CFAR \cite{han_4d_nodate}. This enables real-time applications in automotive scenarios, where detected targets' clouds contain less clutter and more points than traditional CFAR counterparts.

\subsection{Radar Processing Terminology}
\label{sec:terminology}

Before explaining the previous work on generating 4D radar point clouds via detection, we clarify the key terminology used throughout the paper (based on \cite{roldan_deep_2024}) to alleviate any confusion.

\begin{itemize}
    \item \textit{Raw radar data} refers to the complex baseband samples provided by the ADC at each receiver channel \cite{roldan_deep_2024}.
    \item \textit{Radar frame} refers to the set of ADC samples from each virtual channel's Coherent Processing Interval (CPI). It has dimensions of N$_{\textit{fast}}$ × N$_{\textit{slow}}$ × N$_{V \textit{chan}}$, which are, respectively, the number of samples in fast time, the number of samples in slow time and the number of virtual channels \cite{roldan_deep_2024}.
    \item A \textit{radar cube} is the discrete representation of radar data in spherical coordinates in terms of the range, azimuth, elevation, and Doppler estimates that have already been made. Each cell within a \textit{radar cube} contains a scalar value representing the reflected power in that cell \cite{roldan_deep_2024}. It is important to note that each cell does not have the same size, as this depends on the parameters of the specific radar configuration, such as the transmitted bandwidth or antenna array topology. As a result, we expect to see denser cells near the center and sparser cells as we approach the edge of the frame.
    \item An \textit{extended target} occupies multiple cells in one or more dimensions, whereas a \textit{point target} only occupies a single cell. 
    \item A \textit{detection} is the binary value representing whether a \textit{radar cube} cell contains only noise or noise \textit{and} a target \cite{roldan_deep_2024}. This differs from \textit{classification}, which aims to associate a specific class with each detected cell (that is, 'pedestrian','vehicle', 'cone', etc\ldots).
    \item A \textit{3D occupancy grid} is a set of cubes, also in spherical coordinates, that contains ones in voxels that are occupied by detected targets, and zeros otherwise \cite{roldan_deep_2024}. \textit{3D occupancy grids} can be generated from either LiDAR point clouds directly or \textit{radar cubes} through a detector as done in \cite{roldan_deep_2024}. 
    \item \textit{Point cloud} refers to a set of $N_p$ points, each containing $L$ features that result from selecting only those cells containing ones in a \textit{3D occupancy grid} and converting them to Cartesian coordinates \cite{roldan_deep_2024}. We assume that $L = 5$ for radar point clouds as we add the Doppler and power information to the three spatial dimensions, but for LiDAR point clouds, we have no Doppler information, so $L = 4$ is assumed.
\end{itemize}

\subsection{RaDelft Dataset}
\label{sec:radelft}

Although there are many datasets containing LiDAR point cloud or camera examples, few include the raw radar ADC data signals necessary for DL applications for 4D radar processing \cite{brodeski_deep_2019}, and especially few contain complex real-time road scenarios suitable for AD testing. Some prior work has attempted to generate such datasets synthetically using the physical model and calibration data of the radar itself, but this is computationally expensive and unreliable for automotive applications, as the produced data are not indicative of real-world noise \cite{brodeski_deep_2019}. Other works have generated low-scale datasets not meant for testing in AD applications and require precisely labeled LiDAR ground truth labels \cite{cheng_novel_2022}. 

In our work, we use the \textit{RaDelft} dataset from \cite{roldan_see_2024} to train and evaluate our models against the state-of-the-art (SOTA) NN radar detector from \cite{roldan_deep_2024}. The dataset was collected using the demonstration vehicle from \cite{palffy_multi-class_2022} with an additional Texas Instruments MMWCASRF-EVM 4D Radar unit\footnote{https://www.ti.com/lit/ug/tiduen5a/tiduen5a.pdf} mounted 1.5m above the ground and next to a RoboSense Ruby Plus LiDAR (128-layer rotating LiDAR\footnote{https://www.robosense.ai/en/IncrementalComponents/RubyPlus}) \cite{roldan_see_2024}. Data were collected driving through multiple real-world scenarios in Delft, The Netherlands, including suburban and urban environments. Uniquely, this dataset includes all of the raw radar ADC signals, allowing for novel data processing, but our work will rely on their proposed processing pipeline.

The dataset comprises seven 5-minute scenarios resulting in around 21000 radar frames recorded at 10Hz, along with their LiDAR ground truth point clouds. Both sensors were time-synchronized and spatially calibrated to ensure proper alignment between the produced radar frames and LiDAR point clouds, allowing our model to train \textbf{without} needing labeled point clouds. For training and validation, we use scenes 1, 3, 4, 5, and 7 of the dataset, where 10\% of these frames are used for validation. Scenes 2 and 6 are fully utilized for testing. 

\subsection{Prior Work}
\label{sec:related}

\begin{figure}
    \centering
    \includegraphics[width=0.85\linewidth]{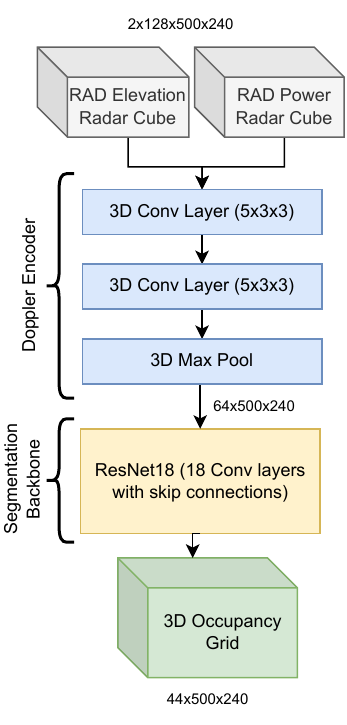}
    \caption{Original proposed NN Radar Detector from \cite{roldan_see_2024} including a Doppler encoder and ResNet18 as the 2D CNN segmentation backbone.}
    \label{fig:original-delft-nn}
    \vspace{-5mm}
\end{figure}

The initial neural network proposed in \cite{roldan_see_2024} and trained with the \textit{RaDelft} dataset consisted of two main components: a Doppler encoder and a 2D Convolutional Neural Network (CNN) segmentation backbone. First, they have two range-azimuth-Doppler radar cubes with dimensions R x A x D where $R = 500$, $A = 240$, and $D = 128$. One cube contains information on the power of each voxel, while the other encodes information about elevation. Due to the lack of Doppler information in LiDAR PCs, it is important to process the Doppler information before estimating the 3D occupancy grid. For this reason, they pass these cubes through a Doppler encoder to preserve the dependency between the Doppler and the angle of the moving targets for improved angular resolution \cite{richards_principles_2010}\cite{roldan_see_2024}. The network's final output is a 3D occupancy grid that can then be used to generate our PC.

\begin{figure}
    \centering
    \includegraphics[width=1.0\linewidth]{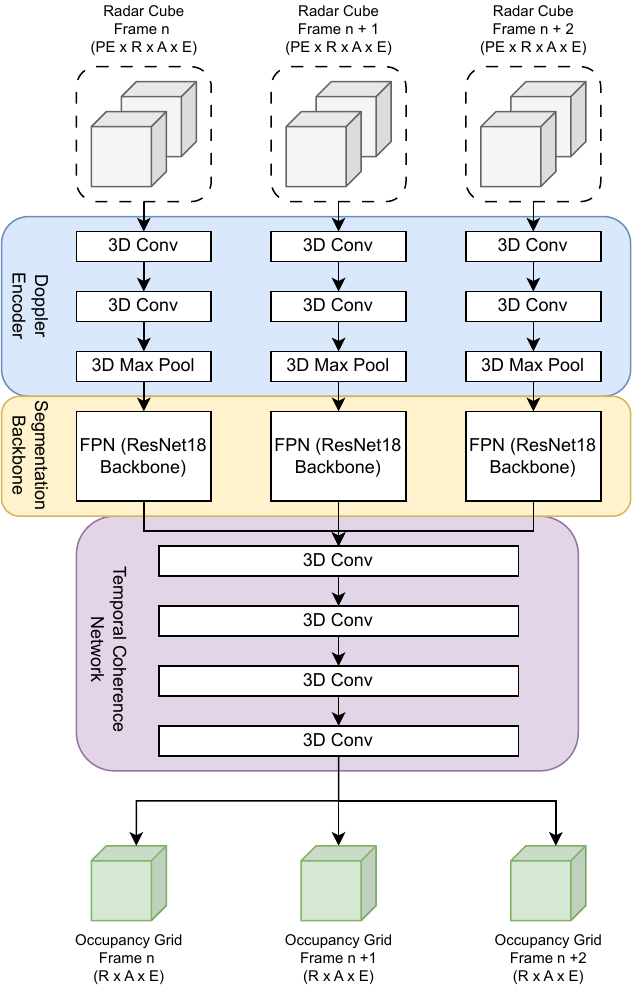}
    \caption{Updated NN Radar Detector architecture from \cite{roldan_deep_2024} with added temporal coherence network.}
    \label{fig:temporal-delft-nn}
    \vspace{-5mm}
\end{figure}

A follow-up ablation study was then performed to determine if temporal information could help minimize BCD and address the "flickering" issue present in radar PCs \cite{roldan_deep_2024}. For example, it is possible to see high instantaneous noise that passes the detection threshold for a single frame and then disappears in subsequent frames, causing 'flickering'. They effectively reduced flickering and smoothed out random noise by taking three data frames and processing them simultaneously. The resulting architecture is similar to the original network seen in \autoref{fig:original-delft-nn}, but with three simultaneous predictions fed into a \textit{temporal coherence network} as shown in \autoref{fig:temporal-delft-nn}. These modifications resulted in a reduction in BCD of ~28\% compared to their original proposed architecture, showing the significant benefit of incorporating temporal information, even with such a simple and efficient network.

%%%%%%%%%%%%%%%%%%%%%%%%%%%%%%%%%%%% METHODS %%%%%%%%%%%%%%%%%%%%%%%%%%%%%%%%%%%%

\section{Methods}
\label{sec:methods}

Our work expands upon that done in \cite{roldan_deep_2024} by investigating the impact of higher-capacity segmentation backbones on the accuracy of our generated point clouds. To this end, we use the same overall architecture from \cite{roldan_deep_2024} and the Focal loss function from \cite{lin_focal_2017}:
\begin{equation}
    FL(p_t) = -\alpha_t(1-p_t)^\gamma log(p_t)
\end{equation}
with
\begin{equation}
    p_t =
    \begin{cases}
        p & \text{if } y = 1 \\
        1-p & \text{otherwise}
    \end{cases}
\end{equation}
where $y \in \{\pm1\}$ is the ground truth class that tells us whether there is detection or not, $\alpha_t$ is the weighting factor where $\alpha \in [0,1]$ for class "1" and $1 - \alpha$ for class "-1", and $\gamma > 1$ is the focusing factor \cite{lin_focal_2017}\cite{roldan_deep_2024}; we use $\alpha = 0.99$ and $\gamma = 2$ when calculating focal loss in training.

In addition, we modify various training hyperparameters and the segmentation backbone. For example, studies have shown that models such as ResNet50 and ResNet101 are highly capable in panoptic segmentation tasks while remaining relatively small (25.6 M and 44.7M parameters, respectively) \cite{elharrouss_backbones-review_2024}\cite{he_deep_2015}. Thus, we aim to explore the effect of swapping the ResNet18 backbone for these models in addition to an extensive model such as ResNet152 (60.2M parameters). Additionally, we further examine the temporal coherence network's impact by evaluating performance with varying 3D convolutional layers in the temporal section of the architecture (0, 2, 4, and 6 layers). In doing so, we aim to better grasp whether or not higher-capacity models can improve the quality of 3D point clouds generated with 4D radar.

\begin{figure}
    \centering
    \includegraphics[width=1\linewidth]{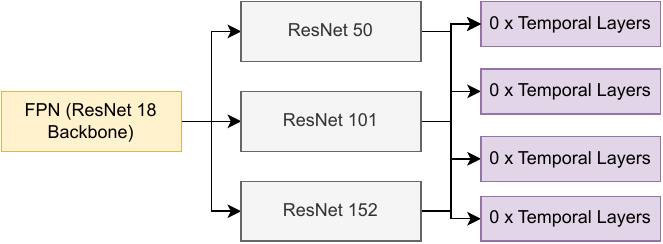}
    \caption{Breakdown of trained models experimented with and compared.}
    \label{fig:seg-models}
    \vspace{-5mm}
\end{figure}
\begin{figure*}[ht]
\vspace{0mm}
    \centering
   \subfloat[LiDAR (Ground Truth)]{
        \includegraphics[width=0.31\linewidth]{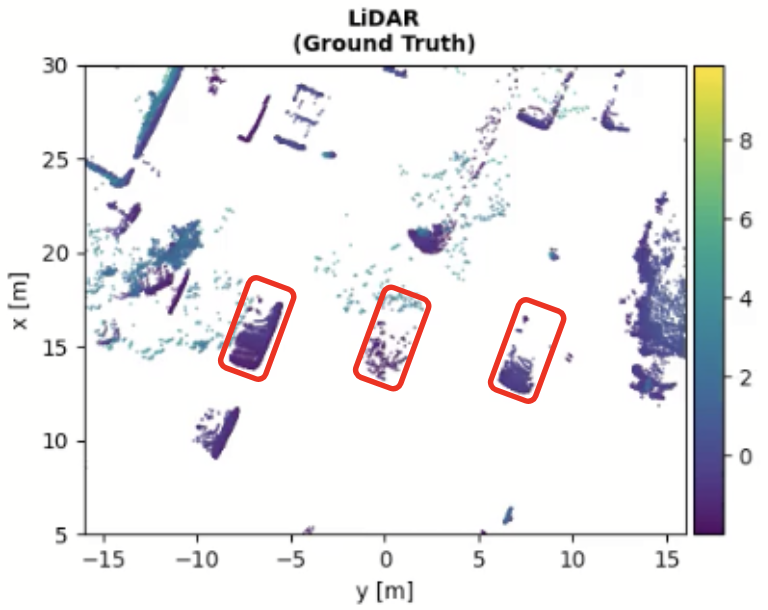}}
    \subfloat[Model Prediction with ResNet18 Backbone]{
        \includegraphics[width=0.31\linewidth]{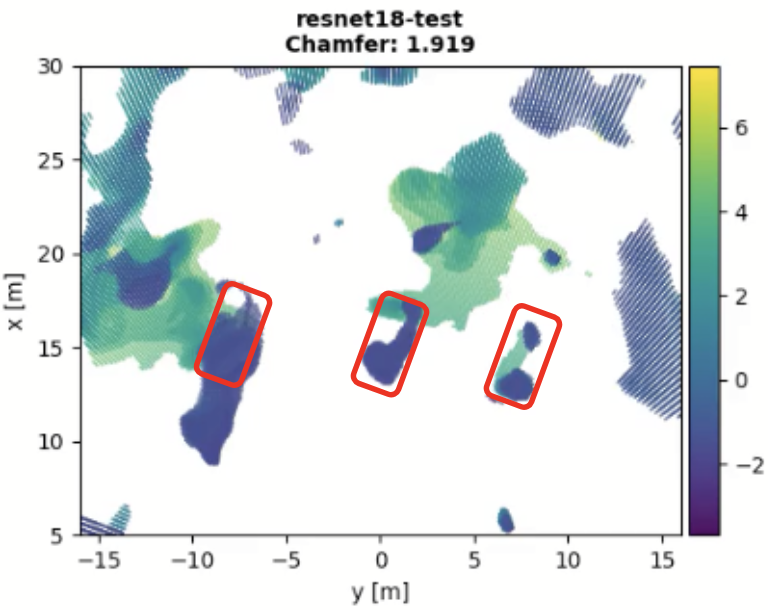}}
    \subfloat[Model Prediction with ResNet152 Backbone]{
        \includegraphics[width=0.31\linewidth]{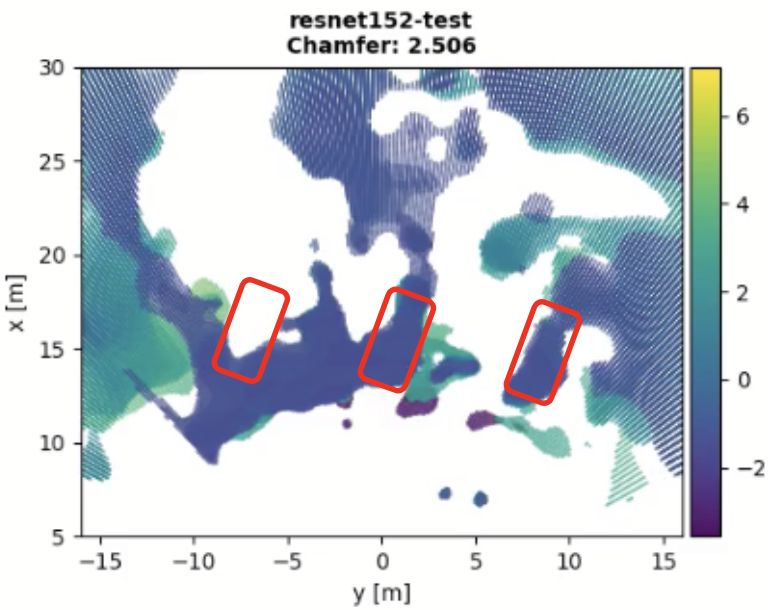}}
    \caption{{\textbf(Experiment: 0 Temporal Layers)} Comparison of model point cloud predictions with ResNet18 and ResNet152 segmentation backbones. Observe the learned noise present with ResNet152, making it difficult to discern detections that are visible in the LiDAR and ResNet18 point clouds.}
   \label{fig:resnet18-vs-resnet152-comparison}
   \vspace{-3mm}
\end{figure*}
We quantitatively evaluate our models using three primary metrics: $P_d$ (probability of detection), $P_{fa}$ (probability of false alarm), and $BCD$ (bidirectional chamfer distance). All three are calculated using LiDAR, $S_1$, as reference, but we use the occupancy grid format for $P_d$ and $P_{fa}$ calculations and the point clouds format for $BCD$. We calculate $P_d$ and $P_{fa}$ using the TP (true positive), FP (false positive), and FN (false negative) values that compare the values in each voxel between the occupancy grids as follows: %\ref{eq:p_d} and \ref{eq:p_fa}.

\begin{equation}
\label{eq:p_d}
    P_d = \frac{TP}{TP + FN}
\end{equation}

\begin{equation}
\label{eq:p_fa}
    P_{fa} = \frac{FP}{|S_1| - TP - FN}
\end{equation}

\begin{figure}[!h]
    \centering
    \includegraphics[width=1\linewidth]{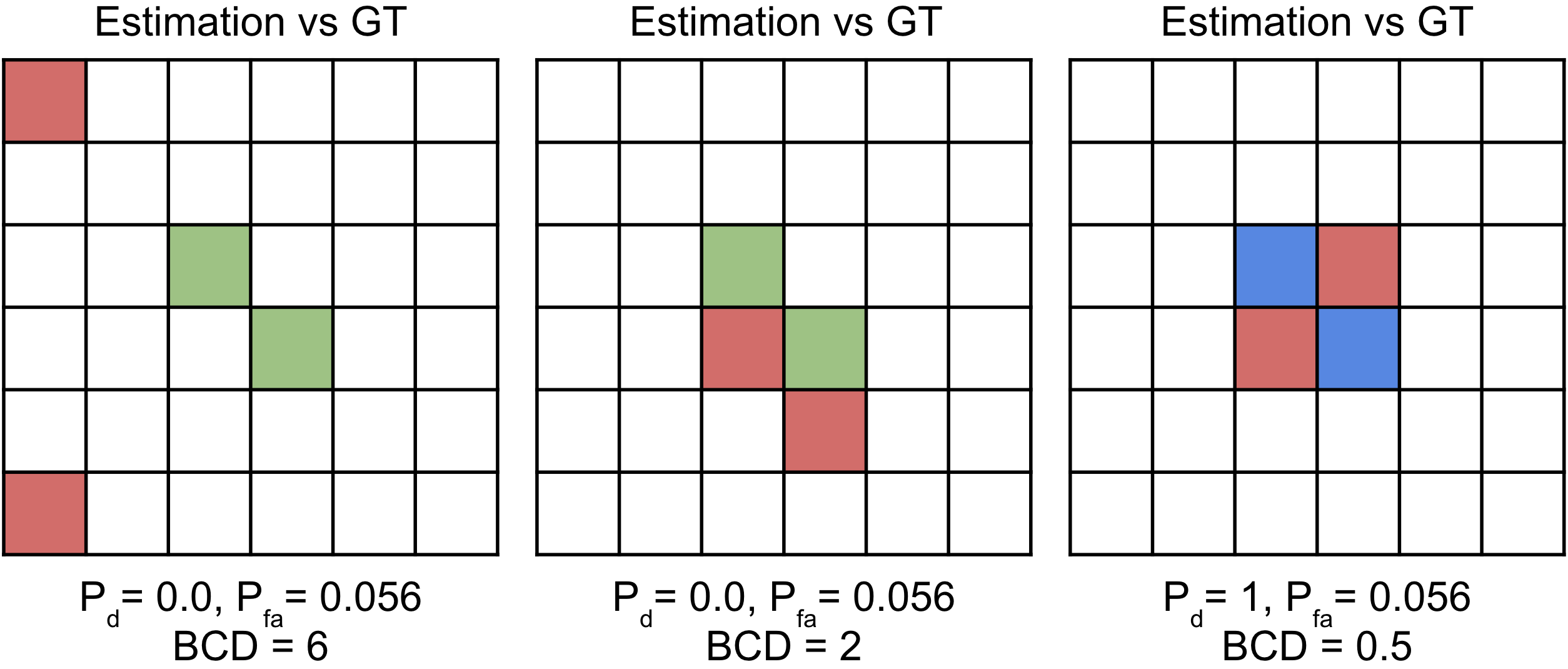}
    \caption{Visual comparison of point cloud prediction metrics. Green represents the ground truth (GT), red is the estimation, and blue is the overlap between the estimation and the GT.}
    \label{fig:metrics-visual-comparison}
    \vspace{-3mm}
\end{figure}

However, one must exercise caution when analyzing these metrics. To illustrate this, suppose there is a misalignment in mounting the LiDAR and 4D Radar sensors of a few centimeters or some small angle that is not corrected for in the signal processing pipeline to correctly align predicted and ground-truth PCs. Such a misalignment causes $P_d$ to drop dramatically while $P_{fa}$ increases, as demonstrated in \autoref{fig:metrics-visual-comparison}. Therefore, a better metric for comparison is BCD, which evaluates the average Euclidean distance between the points of two different point clouds. Following the same example in \autoref{fig:metrics-visual-comparison}, we see in the center plot that when the predicted point cloud is slightly shifted, $P_d=0$, which is the same as the left example where the predicted point cloud is far off from the ground truth. However, when we examine the calculated BCD, we find that we have a better representation of the prediction accuracy, with BCD showing 6 and 2 between the left and center examples, respectively.  Moving forward, there are multiple definitions for the chamfer distance; we specifically follow the calculation method explained in \cite{roldan_deep_2024}, which is as follows:
\begin{equation}
    \begin{split}
        CD (S_1, S_2) = \frac{1}{|S_1|} \sum_{x\in S_1} \min_{y \in S_2} ||x - y||_2 +\\
        \frac{1}{|S_2|} \sum_{y\in S_2} \min_{x \in S_1} ||x - y||_2,
    \end{split}
\end{equation}
where $S_1$ and $S_2$ represent the ground truth LiDAR reference PC and the predicted 4D radar PC to be compared, respectively, and {|S|} is the cardinality of the set \cite{roldan_deep_2024}. In this case, our primary metric for comparison is the minimization of BCD, as this represents the close alignment of our predicted point cloud against the ground truth.

Next, we note that, while we borrow the same overall architecture for the Doppler encoder and other sections of the network, we must alter how we train with specific segmentation models such as ResNet101 and ResNet152. The models are significantly larger than the ResNet18 backbone used initially and, as such, require longer to train than the original 20 epochs. Therefore, we train these larger models for at least 40 epochs in addition to tuning other hyperparameters such as the optimizer, scheduler, and their respective parameters.

However, Additional layers and size do not automatically correlate to better performance. With a larger size, the model is more susceptible to learning noise, resulting in less accurate and incoherent predictions. We can circumvent this issue somewhat by incorporating additional L1 and L2 regularizers to prevent the network using larger backbone models from generalizing to noise during training. Furthermore, we increase the effective batch size from 4 to 8 to allow the model to see more examples before each training step. One last modification is the change from batch normalization in the Doppler encoder to group normalization to help account for the larger batch sizes. With all these changes, we trained the network using ResNet 50, 101, and 152 model backbones with varying combinations of 0, 2, 4, and 6 temporal layers.

%%%%%%%%%%%%%%%%%%%%%%%%%%%%%%%%%%%% RESULTS %%%%%%%%%%%%%%%%%%%%%%%%%%%%%%%%%%%%

\section{Results}
\label{sec:results}

\begin{figure}[h]
    \centering
    \includegraphics[width=1\linewidth]{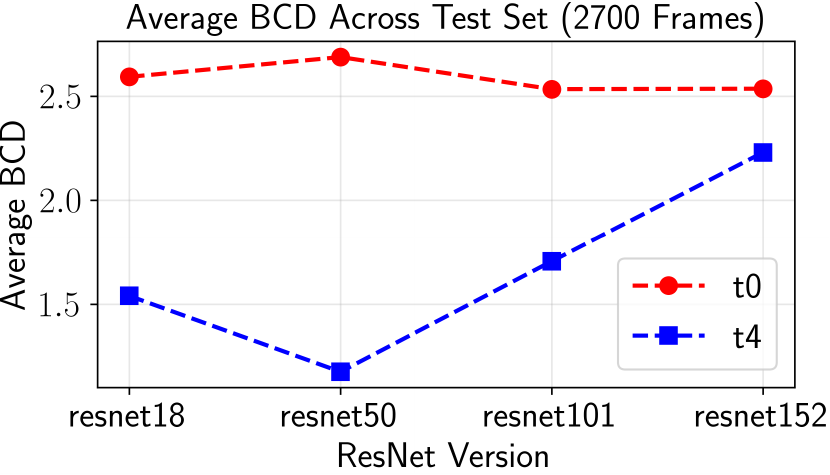}
    \caption{Average BCD over 2700 frames (entire test set) representing key results from \autoref{tab:model_comparison}.}
    \label{fig:model-bcd-plots-avg-2700}
    \vspace{-3mm}
\end{figure}

\begin{table*}[!ht]
\centering
\caption{Comparison of Model Training Results on Test set. The "Inconclusive" denotation represents tests that were unable to meet the 50\% confidence threshold for detections, resulting in no predicted points and near-empty point clouds.}
\begin{tabularx}{0.7\textwidth}{lCCC}
\hline
\textbf{Model} & \textbf{P$_d$(\%)} & \textbf{P$_{fa}$(\%)} & \textbf{Chamfer Distance($m$)} \\
\hline
\textbf{0 Temporal Layers} \\
\hline
ResNet18 \cite{roldan_see_2024} & 58.1     & 6.8     & 2.593 \\
ResNet50            & 56.9     & 7.4     & 2.688 \\
ResNet101           & 61.6     & 7.6     & 2.534 \\
ResNet152           & 61.2     & 7.6     & 2.536 \\
\hline
\textbf{2 Temporal Layers} \\
\hline
ResNet18            & Inconclusive     & Inconclusive     & Inconclusive \\
ResNet50            & 18.6     & 4.5     & 2.101 \\
ResNet101           & 16.1     & 4.1     & 1.820 \\
ResNet152           & Inconclusive     & Inconclusive     & Inconclusive \\
\hline
\textbf{4 Temporal Layers} \\
\hline
\textcolor{RedOrange}{ResNet18 (baseline)} \cite{roldan_deep_2024} & \textcolor{RedOrange}{60.04} & \textcolor{RedOrange}{4.3} & \textcolor{RedOrange}{1.54} \\
\textcolor{ForestGreen}{\textbf{ResNet50}} & \textcolor{ForestGreen}{\textbf{64.8}} & \textcolor{ForestGreen}{\textbf{1.8}} &      \textcolor{ForestGreen}{\textbf{1.175}} \\
ResNet101           & 58.7     & 4.3     & 1.706 \\
ResNet152           & 60.3     & 6.4     & 2.230 \\
\hline
\end{tabularx}%
\label{tab:model_comparison}
\vspace{-5mm}
\end{table*}
%As our current work stands, we find some unique perspectives on the impact segmentation models %have on the overall quality of predicted point clouds.
Through training the network, we find two key results reflected in \autoref{fig:model-bcd-plots-avg-2700}. First, we find that the models with zero temporal layers show an interesting trend: the average BCD over the test set demonstrates a marginal degradation compared to the baseline for the ResNet50 model and an improvement for those trained with ResNet101 and ResNet152. In contrast, when training the network with four temporal layers, we observe that the BCD drastically improves with ResNet50 by around 23.7\% and noticeably degrades with ResNet101 and ResNet152.   

Our general results can be seen in \autoref{tab:model_comparison}, where we split our experiments between tests with varying configurations of temporal layers. We further distinguish our results between two separate categories.

\subsection{Impact of Segmentation Backbones Without the Inclusion of Temporal Data}
\label{sec:results-no-temporal}

While testing various forms of ResNet with zero temporal layers\footnote{0 temporal layers means that we do not process the three consecutive frames proposed in \cite{roldan_deep_2024}, which is also referred to as the "single-time" example and is derived from the base architecture proposed in \cite{roldan_see_2024}.} we see that higher-capacity models, such as ResNet101 and ResNet152, improve the overall chamfer distance by 2.275\%. However, the ResNet50 model sees a reduction in overall performance across $P_d$, $P_{fa}$, and $BCD$, with the chamfer distance increasing by 3.664\%. These results are a marginal difference over our baseline model using ResNet18 and demonstrate that the additional layers found in the higher-capacity segmentation models do not necessarily correlate with a direct improvement in $BCD$. This contradicts our initial intuition that a higher-capacity segmentation backbone will allow for more accurate detections with stronger segmentation capabilities. Still, as we train, we can see that these models are highly susceptible to overfitting or generalizing to noise in the raw radar samples. Without the addition of temporal data, these spikes in instantaneous noise cannot be filtered out, which we can see when we compare the produced point clouds between our models trained with ResNet18 and ResNet152 in \autoref{fig:resnet18-vs-resnet152-comparison}. Therefore, while we may see a reduction in BCD, indicating a higher overlap between point clouds, the additional noise introduced presents significant problems for downstream tasks such as perception or SLAM. 

% \begin{figure}
%     \centering
%     \includegraphics[width=1\linewidth]{images/18_vs_152_0_temporal.png}
%     \caption{Comparison of predicted pointclouds between ResNet18 and ResNet152 with no temporal smoothening.}
%     \label{fig:non-temporal-comparison}
% \end{figure}

\subsection{Impact of Segmentation Backbone and Temporal Smoothing}
\label{sec:results-temporal}
\begin{figure}[!h]
    \centering
    \includegraphics[width=1\linewidth]{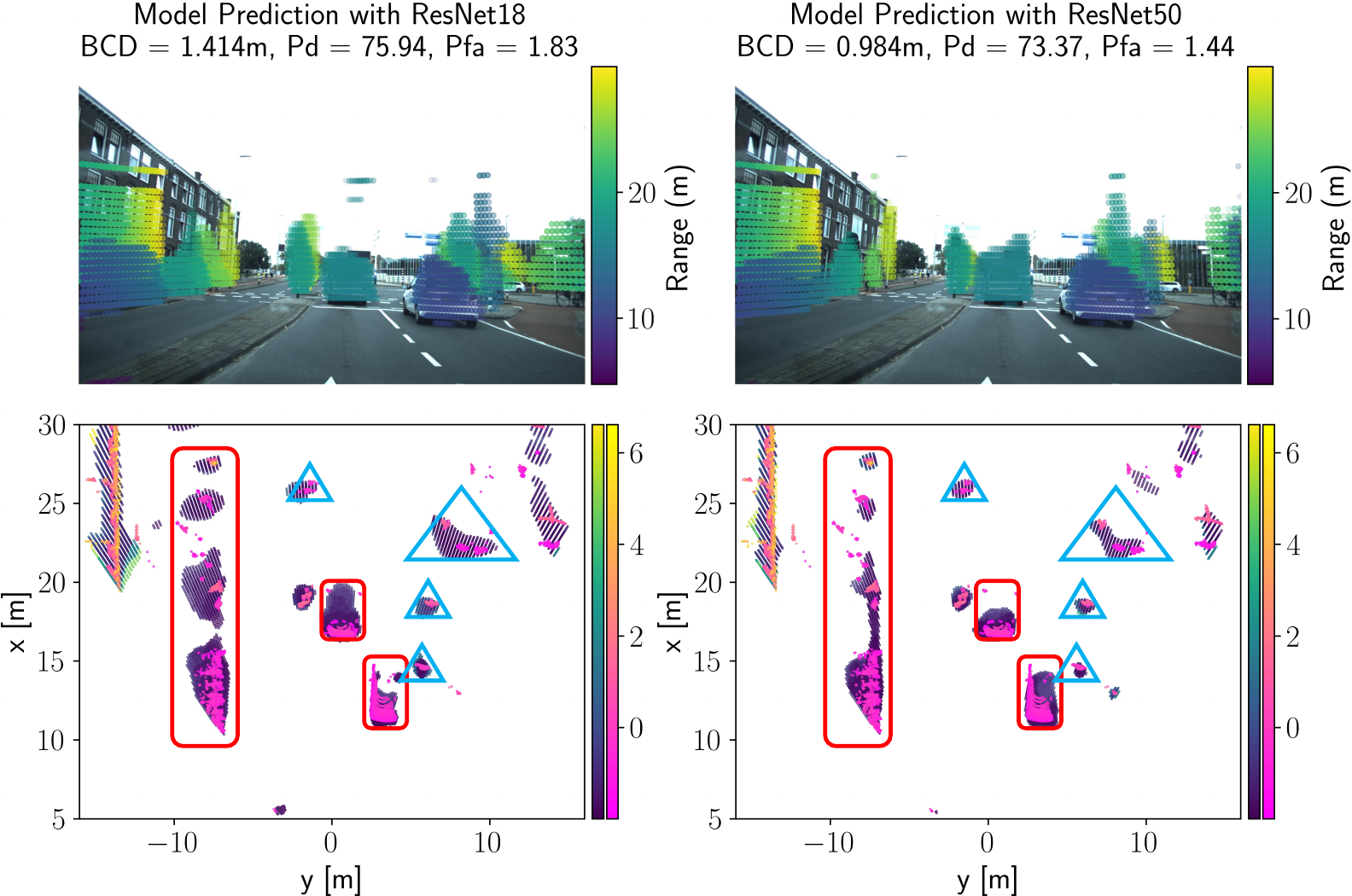}
    \caption{Comparison of the model trained with ResNet50 and four temporal layers (best model) against ResNet18 (baseline) with the LiDAR GT overlayed. We qualitatively observe that the model with ResNet50 has a predicted point cloud that more closely aligns with the GT than the model trained with ResNet18. This is supported quantitatively by the single-frame metrics where we witness a drastic improvement of ~30\% for BCD while $P_d$ and $P_{fa}$ remain fairly similar.}
    \label{fig:killer-example-1}
    \vspace{-5mm}
\end{figure}

\begin{figure*}[ht]
\vspace{1mm}
    \centering
   \subfloat[ResNet18 (Baseline)]{
        \includegraphics[width=0.24\linewidth]{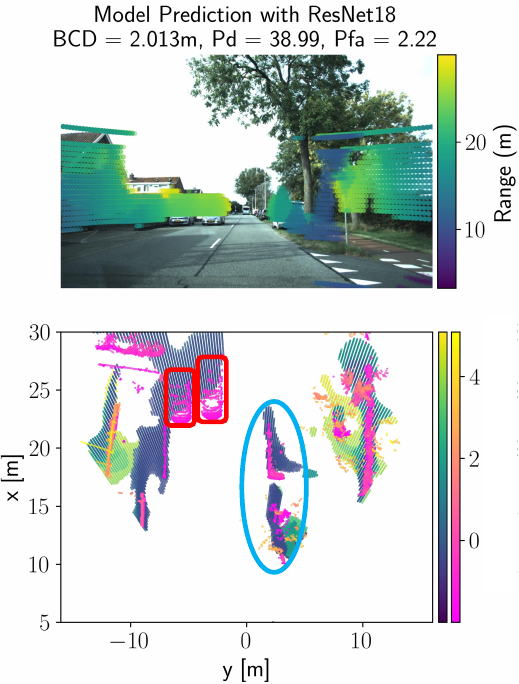}}
    \subfloat[ResNet50]{
        \includegraphics[width=0.24\linewidth]{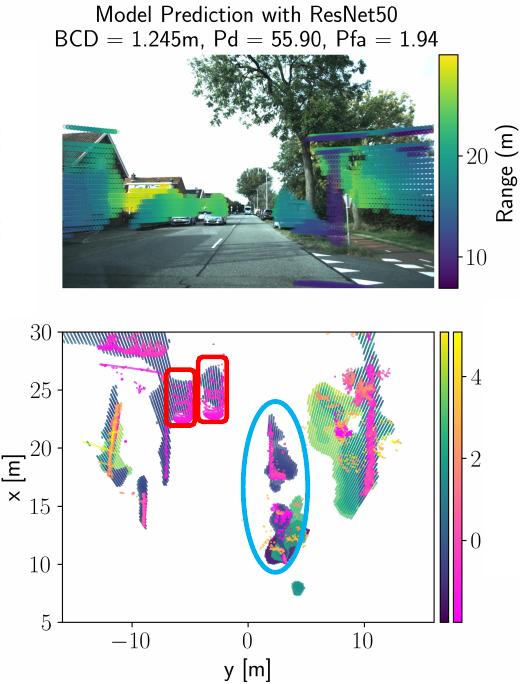}}
    \subfloat[ResNet101]{
        \includegraphics[width=0.24\linewidth]{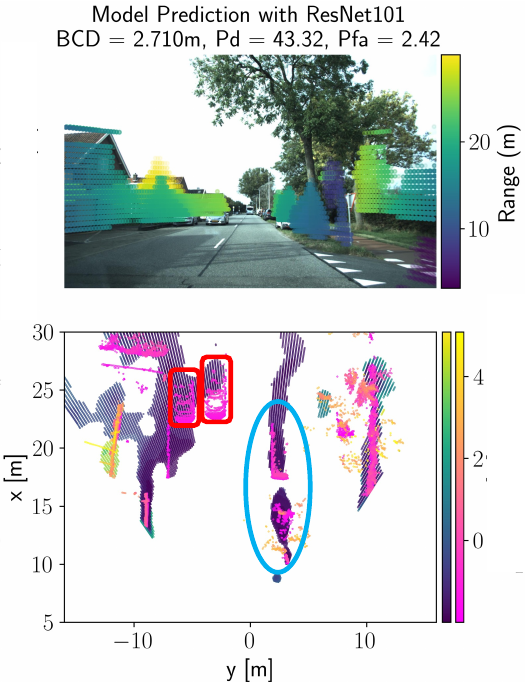}}
    \subfloat[ResNet152]{
        \includegraphics[width=0.24\linewidth]{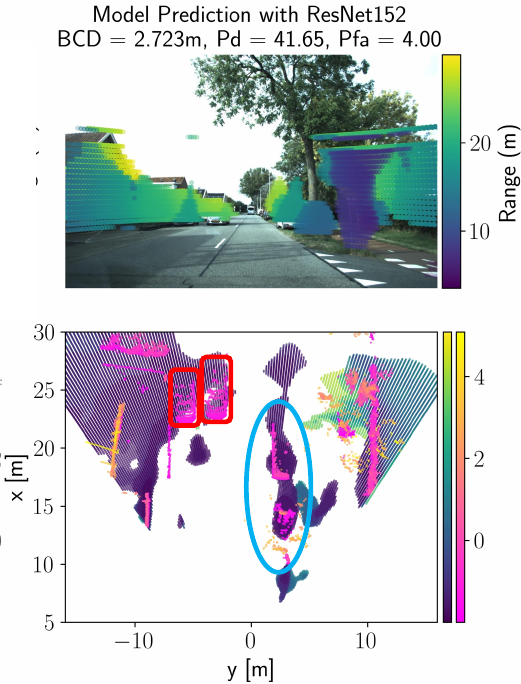}}
        \vspace{-1mm}
    \caption{\textbf{(Experiment: 4 Temporal Layers)} Comparison of model PC predictions across all 4 segmentation backbones. There is an evident improvement when using ResNet50, but ResNet101 and ResNet152 reveal a regression in performance.}
   \label{fig:all-model-visual-comparison}
   \vspace{-6mm}
\end{figure*}
Interestingly, we found that while testing with varying degrees of temporal smoothing, four layers remained the most stable and consistent in training with a window of three frames. Here, we see a noticeable valley in performance as we progress to higher-capacity segmentation models as seen in \autoref{fig:model-bcd-plots-avg-2700}. In particular, ResNet50 shows noticeable improvements in BCD over ResNet18 at roughly 23.7\%. This improvement in BCD is corroborated by an improvement in both $P_d$ and $P_{fa}$ as seen in \autoref{tab:model_comparison}. As we introduce temporal smoothing, slight improvements in point prediction accuracy from higher-capacity models are likely to be compounded to produce these drastic improvements, along with small fine-tuning of the training hyperparameters.

This compounding factor explains the performance discrepancy between ResNet50 and the larger models. Improvements in prediction accuracy are overcome by the inability to completely filter out the additional noise learned by these larger models. This also explains why ResNet101 and ResNet152 perform poorly in terms of BCD compared to the baseline. 

Furthermore, when qualitatively analyzing the results, we see that the lack of additional noise learned with ResNet50, along with slight improvements in prediction accuracy, leads ResNet50 to show excellent qualitative results, as seen in \autoref{fig:killer-example-1} where the predicted point clouds are sharper and more accurately aligned with the GT. Observing models trained with ResNet101 and ResNet152, it is clear that these models are not necessarily overfit to the training set when qualitatively viewing the predictions between the training, test, and validation sets and quantitatively observing a continuous decline in train and validation loss during training. A more reasonable explanation is that with the larger-capacity model, we are more susceptible to generalizing to noise from the abundant input radar signals, which we saw when comparing the predicted PCs between the train and test sets.

% \subsubsection{Two and Six Temporal Layers}
% \label{sec:two-and-six-temporal-layers}

Additionally, we note that two temporal layers did offer an improvement over the single-time (zero-layer) example but inconsistently produced PCs with confidence past the fifty-percent threshold, resulting in nearly "empty" PCs, thus the inconclusive denotation in \autoref{tab:model_comparison}. 

Finally, we find that no model can produce points with confidence greater than fifty percent at six temporal layers, resulting in empty point clouds. The temporal layers assist in erasing the instantaneous noise that occurs from frame to frame, so with six layers, we see the model effectively smooth out or erase all of the points in the scene as they are learned as instantaneous "noise." Hence, the lack of inclusion of results for the six-layer models in \autoref{tab:model_comparison}. 

% \begin{figure*}[ht]
%     \centering
%     \includegraphics[width=1\linewidth]{images/comprehensive-comparison/ex1.png}
%     \caption{{\textbf(Experiment: 4 Temporal Layers)} Comparison of model point cloud predictions across all 4 segmentation backbones. Observe how we see improvements over baseline when using ResNet50, but ResNet101 and ResNet152 reveal a regression in performance over baseline.}
%     \label{fig:all-models-visual-comparison}
%     \vspace{-6mm}
% \end{figure*}

%%%%%%%%%%%%%%%%%%%%%%%%%%%%%%%%%%%% CONCLUSION + FUTURE WORKS %%%%%%%%%%%%%%%%%%%%%%%%%%%%%%%%%%%%

\section{Conclusion}
\label{sec:conclusion}

In this paper, we examined the effects of various segmentation backbones and degrees of temporal smoothing on the overall performance of 4D radar point cloud predictions. Four temporal layers were found to be optimal, but medium-sized residual network models such as ResNet50 remain the most optimal solution in terms of the bidirectional Chamfer distance, $P_d$, and $P_{fa}$, where we found an improvement of more than 24\% in BCD over baseline, as shown in \autoref{fig:model-bcd-plots-avg-2700} and \autoref{tab:model_comparison}. However, contrary to what one might expect, larger ResNet models like 101 and 152 may harm performance due to their susceptibility to overgeneralization to noise, making indiscernible point cloud detections that are unusable for Autonomous Driving applications. 

% \subsection{Future Works}
% \label{sec:future-works}

While working with the \textit{RaDelft} dataset, we found many opportunities for future improvements. For example, acquiring our own dataset filled with more diverse scenes and US roads for testing and research purposes has great potential. One main advantage of Radar is its prowess in inclement weather conditions, but the RaDelft dataset does not provide such scenarios. It is possible that collecting data in Pittsburgh and other cities in the US will allow us to capture more unique changes in weather, elevation, wildlife, and traffic patterns while being one of the few datasets that includes raw 4D ADC samples in an automotive setting, similar to \cite{palffy_multi-class_2022} and \cite{roldan_see_2024}.

Additionally, research has trended toward sequential 4D point cloud processing, where we retain our 3D representation but process with an additional time dimension or encoding of the spatiotemporal relationship across frames \cite{wang_sequential_2022}\cite{fei_comprehensive_2022}\cite{gao_ramp-cnn_2021}. We have seen the benefit of temporal information in minimizing the chamfer distance in this work by focusing on the detector architecture, but this can be taken even further with methods found in \cite{wang_semantic-supervised_2025} and \cite{yin_lidar-based_2020}, with the added potential to improve $P_d$ and $P_{fa}$. By combining concepts learned from the detection and reconstruction works in 4D radar processing, it is possible that we can obtain significant improvements in predicted point cloud accuracy while also addressing the issues of point sparsity and resolution, all while remaining real-time for AD applications.

% \begin{figure}[h]
%     \centering
%     \includegraphics[width=1\linewidth]{images/plots/rolling_bcd_1500frames.png}
%     \caption{per frame rolling average (up to 1500 frames)}
%     \label{fig:model-bcd-plots-per-frame-avg}
%     \vspace{-3mm}
% \end{figure}

% \begin{figure*}[ht]
% \vspace{1mm}
%     \centering
%    \subfloat[per frame rolling average (up to 1500 frames)]{
%         \includegraphics[width=0.48\linewidth]{images/plots/rolling_bcd_1500frames.png}}
%     \subfloat[Overall average over 2700 frames (entire test set) representing data in \autoref{tab:model_comparison}]{
%         \includegraphics[width=0.48\linewidth]{images/plots/average_bcd_2700frames.png}}
%         \vspace{-1mm}
%     \caption{INSERT HERE.}
%    \label{fig:model-performance-plots}
%    \vspace{-6mm}
% \end{figure*}

\printbibliography
\end{document}

%% file: commands.tex
\usepackage{amsmath,amsfonts}
\usepackage{graphicx}
\usepackage{textcomp}
\usepackage{comment}

\usepackage{amsthm}
\usepackage[utf8]{inputenc}
\usepackage[T1]{fontenc}
\usepackage{enumerate}
\usepackage[font=scriptsize,labelformat=simple]{subcaption}

\usepackage{soul}

\usepackage{hyperref}
\hypersetup{
    colorlinks=true,
    citecolor=black,
    linkcolor=black,
    filecolor=black,      
    urlcolor=black,
}

\usepackage[style=ieee]{biblatex}
  
\AtBeginBibliography{\scriptsize}  %\small %\footnotesize %\scriptsize
\usepackage{siunitx}
\usepackage{multirow}
\usepackage{subfig}

\usepackage[ruled,vlined]{algorithm2e}
\DontPrintSemicolon

\usepackage{algpseudocode}

\usepackage{amssymb}    % For additional math symbols
\usepackage{mathtools}  % For more math tools
\usepackage{bm}         % For bold math symbols

\usepackage{accents}  % Add this in the preamble

\usepackage{colortbl}  % For table colors
\usepackage{array}     % For custom column width
\usepackage{adjustbox} % To fit table in text width
\usepackage{wasysym}  % For \lightning and \hourglass
\usepackage{xcolor}  % For colored circles
\usepackage{pifont}% http://ctan.org/pkg/pifont
\definecolor{gold}{HTML}{E5AE1C}
\definecolor{safe}{HTML}{4169E1}
\definecolor{unsafe}{HTML}{C41230}

\theoremstyle{definition}